\documentclass[]{spie}  %

\usepackage{amsmath,amsfonts,amssymb}
\usepackage{graphicx}
\usepackage{multirow}
\usepackage[colorlinks=true, allcolors=blue]{hyperref}
\usepackage{subcaption}
\usepackage{float}

\usepackage{fontawesome}
\usepackage{xcolor}
\newcommand{\greenright}{\textcolor{green}{\faLongArrowRight}}
\newcommand{\reddright}{\textcolor{red}{\faLongArrowRight}}
\title{Learning disentangled representations for explainable chest X-ray classification using Dirichlet VAEs}
\author[1,2]{Rachael Harkness}
\author[1,2]{Alejandro F Frangi}
\author[2,3]{Kieran Zucker}
\author[1,2]{Nishant Ravikumar}
\affil[1]{CISTIB Centre for Computational Imaging and Simulation Technologies in Biomedicine, School of Computing}
\affil[2]{University of Leeds, Leeds, United Kingdom}
\affil[3]{LIMR Leeds Institute of Medical Research}

\begin{document} 
\maketitle

\section{ABSTRACT}

This study explores the use of the Dirichlet Variational Autoencoder (DirVAE) for learning disentangled latent representations of chest X-ray (CXR) images. Our working hypothesis is that distributional sparsity, as facilitated by the Dirichlet prior, will encourage disentangled feature learning for the complex task of multi-label classification of CXR images. The DirVAE is trained using CXR images from the CheXpert database, and the predictive capacity of multi-modal latent representations learned by DirVAE models is investigated through implementation of an auxiliary multi-label classification task, with a view to enforce separation of latent factors according to class-specific features. The predictive performance and explainability of the latent space learned using the DirVAE were quantitatively and qualitatively assessed, respectively, and compared with a standard Gaussian prior-VAE (GVAE). We introduce a new approach for explainable multi-label classification in which we conduct gradient-guided latent traversals for each class of interest. Study findings indicate that the DirVAE is able to disentangle latent factors into class-specific visual features, a property not afforded by the GVAE, and achieve a marginal increase in predictive performance relative to GVAE. We generate visual examples to show that our explainability method, when applied to the trained DirVAE, is able to highlight regions in CXR images that are clinically relevant to the class(es) of interest and additionally, can identify cases where classification relies on spurious feature correlations.

\textbf{Keywords:} Multi-label classification, chest X-ray images, variational autoencoders, Dirichlet distribution, disentanglement, explainability

\textbf{Copyright notice:}
Copyright 2023 Society of PhotoOptical Instrumentation Engineers. One print or electronic copy may be made for personal use only. Systematic reproduction and distribution, duplication of any material in this paper for a fee or for commercial purposes, or modification of the content of the paper are prohibited.

\section{INTRODUCTION}
Medical image interpretation is a complex and challenging task requiring in-depth understanding of anatomy and physiology, and years of education and experience. In particular, the interpretation of chest X-rays (CXRs), a 2D projection of 3D thoracic organs/structures, can be especially challenging. The resulting `overlapping' tissue features makes identification of object boundaries challenging due to insufficient tissue contrast, impeding detection of abnormalities. Additionally, pre-existing conditions and comorbidities are common in patients that require acute chest X-ray exams, or are frequently invited to screenings (e.g. chronic obstructive pulmonary disorder and lung cancer screenings). Consequently, in practice multiple pathologies are often observed in a single exam. Co-occurring disease features can appear similarly and can often create diagnostic uncertainty for both expert radiologists and deep learning systems. The learning of explainable representations of clinically complex medical images, where anatomical and extrinsic image acquisition-related features are disentangled from relevant disease features, is a crucial step in overcoming these challenges.

In this study, we propose to use variational autoencoders (VAEs) to learn disentangled multi-modal latent representations of CXRs for explainable classification of multiple co-occurring labels/class (i.e. multi-label classification) \cite{kingma}. VAEs are able to learn compressed representations of data, in which observed variations of salient visual features are captured by a number of latent factors. Disentanglement is achieved when each factor corresponds uniquely to some meaningful composite part of the data. Disentangled representations offer a more explainable latent space, unlike their entangled counterparts. The matching of latent factors to interpretable variations in the data manifold, afforded by disentangled representations, makes them advantageous to downstream predictive tasks that utilise the learned latent space (e.g. classification, regression, clustering) \cite{recent_advances}. Popular approaches to learning disentangled representations with VAEs, includes, $\beta$-VAE \cite{beta}, Factor-VAE\cite{factorvae} and $\beta$-TCVAE \cite{betatc_vae}.

While these approaches improve disentanglement in Gaussian-prior VAEs, success is limited due to the dense and unimodal nature of a multivariate Gaussian prior. With this in mind, we propose to use a VAE with a Dirichlet prior, referred to as Dirichlet VAE (DirVAE). We hypothesise that a DirVAE will allow a multi-modal latent representation of CXRs to be learned through multi-peak sampling and will encourage latent disentanglement due to the sparse nature of the Dirichlet prior \cite{var_sparse_coding}.

To evaluate the potential benefits of DirVAE over a conventional VAE with a Gaussian prior (GVAE), we use the CheXpert dataset to tackle a complex multi-label classification problem \cite{chexpert}. We convert the multi-label task to a series of binary classification problems. Logistic regression classifiers are trained jointly within a representation learning-centered framework, to predict each class of interest, with all binary classification tasks tackled simultaneously. Specifically, we jointly train the class-specific logistic regression classifiers with the DirVAE, to encourage disentanglement of class-specific latent factors, and evaluate model capacity for learning diagnostically separable representations of CXRs. Model evaluations investigate classifier performance, latent factorisation and its explainability. We train and evaluate the GVAE identically to the DirVAE for fair comparison, and highlight the benefits of the latter in terms of predictive performance and explainability.

\section{METHODOLOGY}

\subsection{Data}
\begin{figure}[ht]
    \centering
    \includegraphics[width=0.63\textwidth]{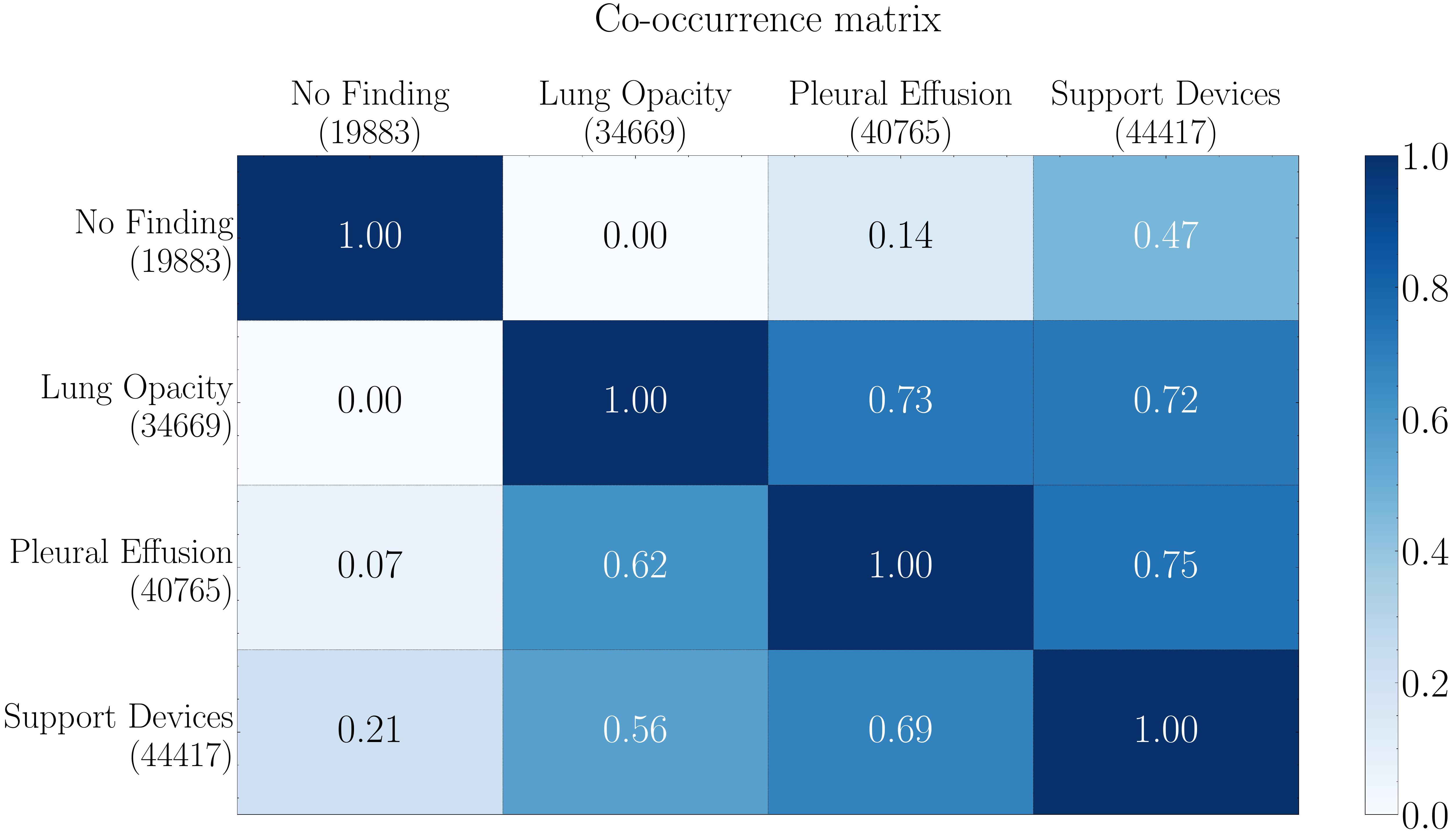}
    \caption{Label co-occurrence matrix of multi-label CheXpert data (training data)}
    \label{fig:cooccurence}
\end{figure}

The CheXpert data set, one of the largest publicly available CXR datasets with multi-label outcomes, is used throughout this study. To overcome issues of severe class imbalance in the data, and simplify training of the logistic regression classifiers, we sample 17,000 images of each of the four most represented classes featured in the CheXpert dataset, namely, `No Finding', `Lung Opacity', `Pleural Effusion', and `Support Devices'. Figure \ref{fig:cooccurence} presents the frequency of classes of interest as well as their frequency of co-occurrence. We split the CheXpert dataset into train $(n=52,943)$, validation $(n=5057)$ and test $(n=10,000)$ sets.

\subsection{Dirichlet VAE} The goal of this study is to explore the benefits of using the DirVAE to learn a multi-modal latent representation of CXR images, comprised of explainable latent factors. To accomplish this we model the Dirichlet prior explicitly as in Gawlikowski, J. et al. (2022). \cite{gawlikowski2022advanced}. 
Alternative approaches frequently utilize a softmax-Laplace approximation \cite{topic_srivastava}. However, such approaches to modelling the Dirichlet prior prevents true approximation of multi-modal representations as shown in Joo, W. et al. (2020) \cite{joo2020dirichlet}. The Dirichlet distribution is a continuous multivariate probability distribution defined over a set of discrete distributions. It is parameterised by a $K$-dimensional vector typically referred to as the concentration, where, $K$ corresponds to the set of discrete distributions. Here, the set of discrete distributions represent the latent space in a DirVAE. 

We define the Dirichlet probability density function as,

\begin{equation}
    \text{Dir}(x;\alpha) = \frac{\Gamma(\sum\alpha_k)}{\prod\Gamma(\alpha_k)} \prod x_k^{\alpha_k-1} , \hspace{1cm} \left( x_k \geq 0; \hspace{0.2cm} \sum\limits_{k}x_k = 1
    \right)
    \label{eq:dir}
\end{equation}

where, $\alpha_k>0$ and $\Gamma(\cdot)$ denotes the gamma function. The Dirichlet distribution can expressed with a composition of gamma random variables with the Gamma distribution defined as, 

\begin{equation}
    \text{Gamma}(x; \alpha, \beta) = \frac{\beta^\alpha}{\Gamma(\alpha)} x^{\alpha-1} e^{-\beta x},
    \hspace{1cm}
    \left( \alpha > 0; \hspace{0.2cm} \beta > 0 \right)
\end{equation}

where $\beta$ is a constant, if $X_k \sim$ Gamma $(\alpha _k)$ are independently distributed, then $(X_1 /X_0, \dots , X_n/X_0)$  follows a Dirichlet distribution parameterised by a vector of $\alpha _1, \dots , \alpha _n$, with $X_0 = \sum X_k$. Alternatively, we can say where $K$ independent random variables are sampled from the multivariate Gamma distribution $X \sim$ MultiGamma$(\alpha, \beta \cdot \mathbf{1}_K)$, then we get $\mathbf{Y} \sim$ Dir$(\alpha)$, where $Y_k = X_k/X_0$ \cite{joo2020dirichlet}.

\begin{figure}[h]
    \centering
    \includegraphics[width=0.65\textwidth]{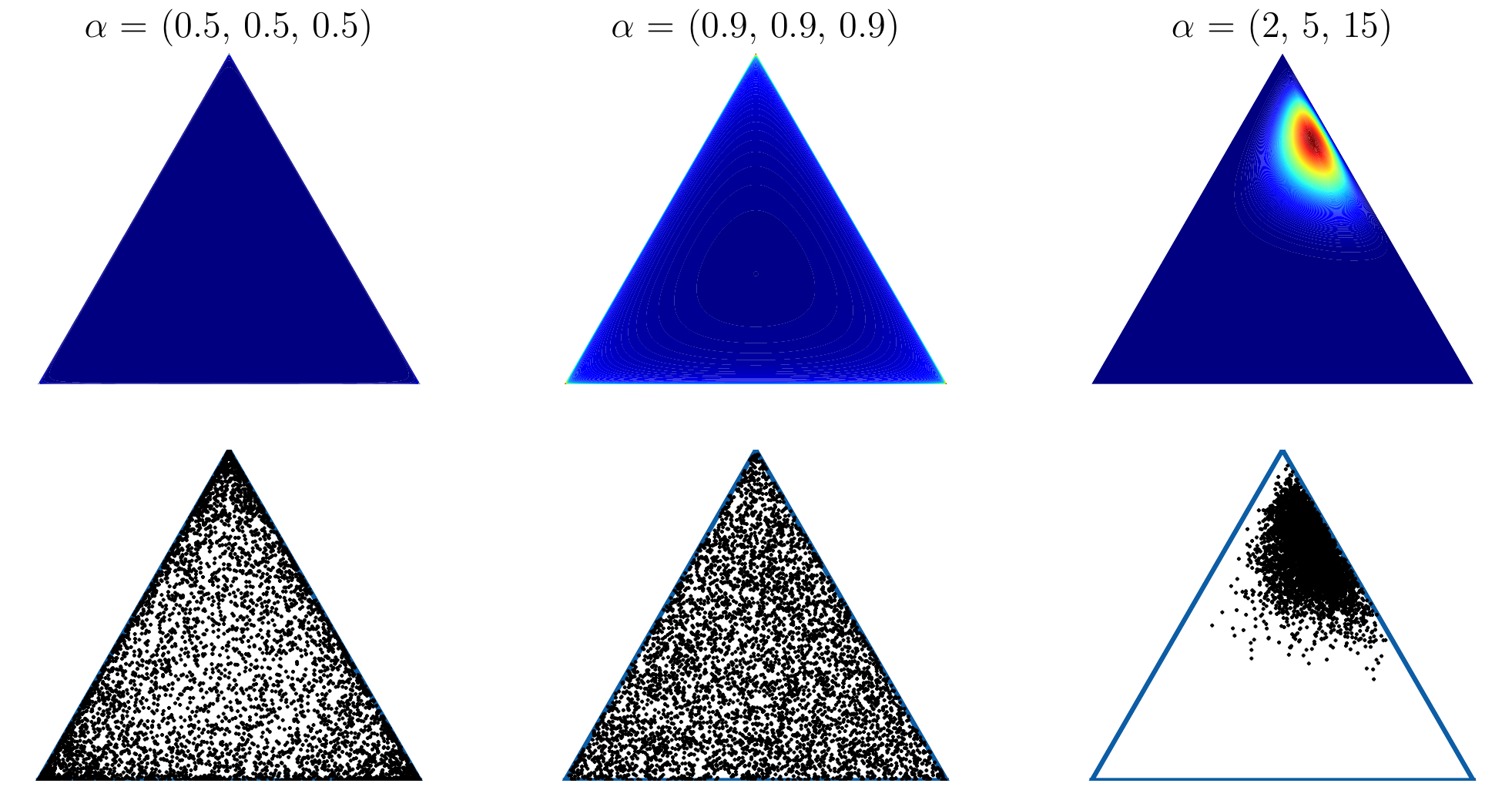}
    \caption{3-dimensional probability density functions from the Dirichlet distribution, each with different $\alpha$ values. The values of $\alpha$ are set to (0.5, 0.5, 0.5), (0.9, 0.9, 0.9) and (2, 5, 15).}
    \label{fig:alphas_dir}
\end{figure}

The Dirichlet concentration governs the shape of distribution on the probability simplex and is selected according to an acceptable trade-off between distributional sparsity and distributional smoothness. This can be observed in Figure \ref{fig:alphas_dir}. For enhancing explainability and to capture multi-modal latent distributions a Dirichlet concentration that enforces sparsity is essential. For these reasons we apply a Dirichlet prior parameterised with a concentration of $0.5 \cdot \mathbf{1}_{1024}$, where $\mathbf{1}_{1024}$ represents the $K$-dimensional latent space with $K=1024$. This latent dimension was used throughout this study for both the DirVAE and the GVAE.

\subsection{Experimental Settings} 

Both the DirVAE and GVAE were trained in four stages categorised as follows: (1) reconstruction, (2) reconstruction and regularisation, (3) classifier initialisation, and (4) joint training. During the first stage, the model is trained using just the reconstruction loss (L1 loss), defined as $L_1 = \sum_{i=1}^{n}| x_i - f(x_i)|$, where $f$ denotes the encoder-decoder functions. In the second stage, the Kullback-Leibler (KL) divergence term is used to regularise model training by minimising the divergence between the approximated posterior distribution of the latent factors and the assumed prior distribution (i.e. Dirichlet for DirVAE and centered multivariate Gaussian for GVAE). KL divergence is defined as $L_{KL}= D_{KL}(q_\phi(z|x) \parallel p_\theta(z|x))$, and for DirVAE is derived as in below,

\begin{equation}
    KL(Q\parallel P) = \sum log \Gamma(\alpha_k) - \sum log \Gamma(\hat{\alpha}_k) + \sum(\hat{\alpha}_k - \alpha_k)\psi(\hat{\alpha}_k)
\end{equation}

where $P$ is the prior distribution, equal to MultiGamma$(\alpha, \beta \cdot 1_k)$, $Q$ represents the learned posterior, $Q = $ MultiGamma$(\hat{\alpha}, \beta \cdot 1_k)$ and $\psi$ is a digamma function \cite{joo2020dirichlet}.

In the next stage, the weights of the pre-trained VAE are frozen and each of the four class-specific logistic regression classifiers are independently optimised for their respective binary classification tasks with binary cross entropy loss, using CXR latent representations as inputs. Once each classifier is fully optimised, the ensemble of logistic regression models are trained together, this objective takes the form,

\begin{equation}
    L_{BCE} = \frac{1}{4} \sum_{c=1}^4 -\frac{1}{n} \sum_{i=1}^{n} (y_{c,i} \cdot log (p_{c,i}) + 1-y_{c,i} \cdot log(1- p_{c,i})),
\end{equation}

where $y$ represents the class labels, $p$ is the predicted probabilities and $c$ represents the different classes.

In the final stage of training, the VAEs and logistic regression classifiers are trained jointly, where, the reconstruction loss, KL divergence loss and all four classifier losses are combined into a single training objective and minimised. Based on preliminary experiments, we found that training the multi-label GVAE and DirVAE models in this stage-wise manner helped stabilise the training process, relative to training the models end-to-end from the beginning. Optimising the latent space with the multi-label classification task encourages latent factors to explain class-specific visual features and allows the classifiers to predict target classes from the learned latent space. This enhances explainability of the learned latent space as latent factors with higher importance/discriminative power for predicting a given class correctly, should correspondingly encode visual features in the CXR image that are representative of that class. 

Our working hypothesis was that this method of training combined with the properties of a DirVAE should yield a disentangled latent space resulting in improved explainability over the GVAE. Both the DirVAE and GVAE were optimized using the Adam algorithm. Logistic regression models were optimized under SGD.

\begin{figure}[H]
    \centering
    \includegraphics[width=0.65\textwidth]{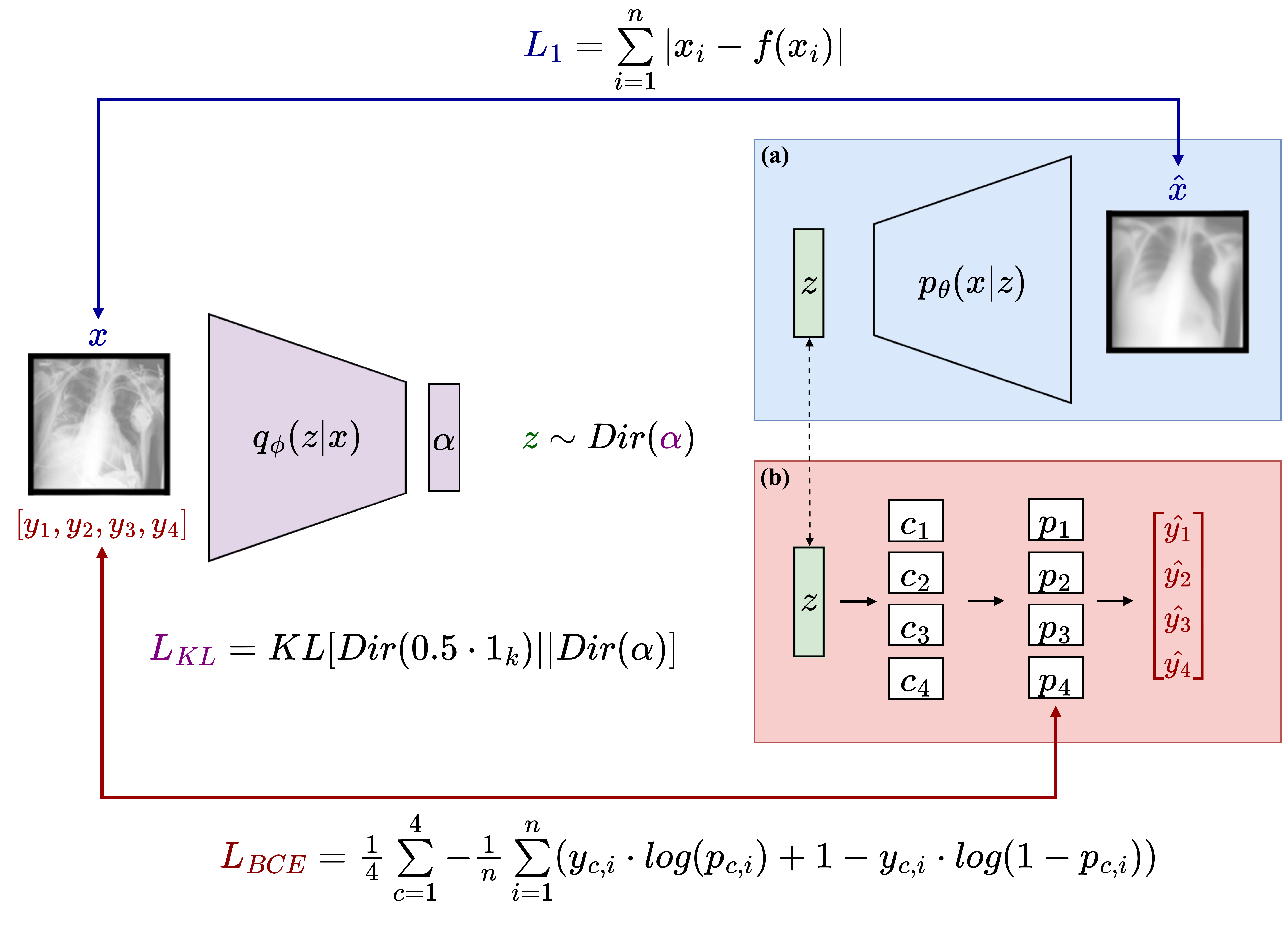} %
    \caption{Dirichlet VAE encoder with a (a) decoder which generates image reconstructions from the latent space $z$ and (b) an ensemble of logistic regression classifiers $C$, each predicting independent class probabilities $p$ from latent space $z$. Classification thresholds are applied to the probabilities to produce class predictions $\hat{y}$.}
    \label{fig:my_label}
\end{figure}

\subsection{Evaluations}

The goal of this study is to evaluate the capacity of the DirVAE to learn explainable, disentangled latent representations of CXR images, for multi-label classification. Accordingly, we assess the performance of the DirVAE by evaluating the performance of their logistic regression classifiers on the multi-label classification problem, and qualitatively assessing the explainability of their respective learned latent spaces. Classifier performance is evaluated through ROC curves and standard classification metrics (evaluated per-class). All results are compared directly with GVAE performance. We perform a qualitative assessment of the explainability of the latent space through a series of `latent factor traversals'. Once the GVAE and DirVAE models were trained, given a CXR image from the test set as input, we first identify a single latent factor with the largest gradient activation from a positive classifier (i.e. where the positive class of interest to a classifier was predicted correctly for the input image). %
The identified latent factor is adjusted incrementally or `traversed', while all other factors are preserved. We visualise the influence of changing the single factor of interest on the decoder reconstructions. Intuitively, if changing only a single latent factor results in class-specific structural changes in the CXR reconstructions, we can assume that the latent space has been successfully disentangled into visual features relating to the class of interest. Pixel-wise variance is created to summarise changes across a traversal and identify features controlled by a specific latent factor. In these evaluations, cases with co-occurring labels are not considered, to simplify interpretation (as co-occurring labels would require exploring multiple relevant latent factors).

\section{RESULTS}
\begin{figure}[H]
    \centering
    \includegraphics[width=\textwidth]{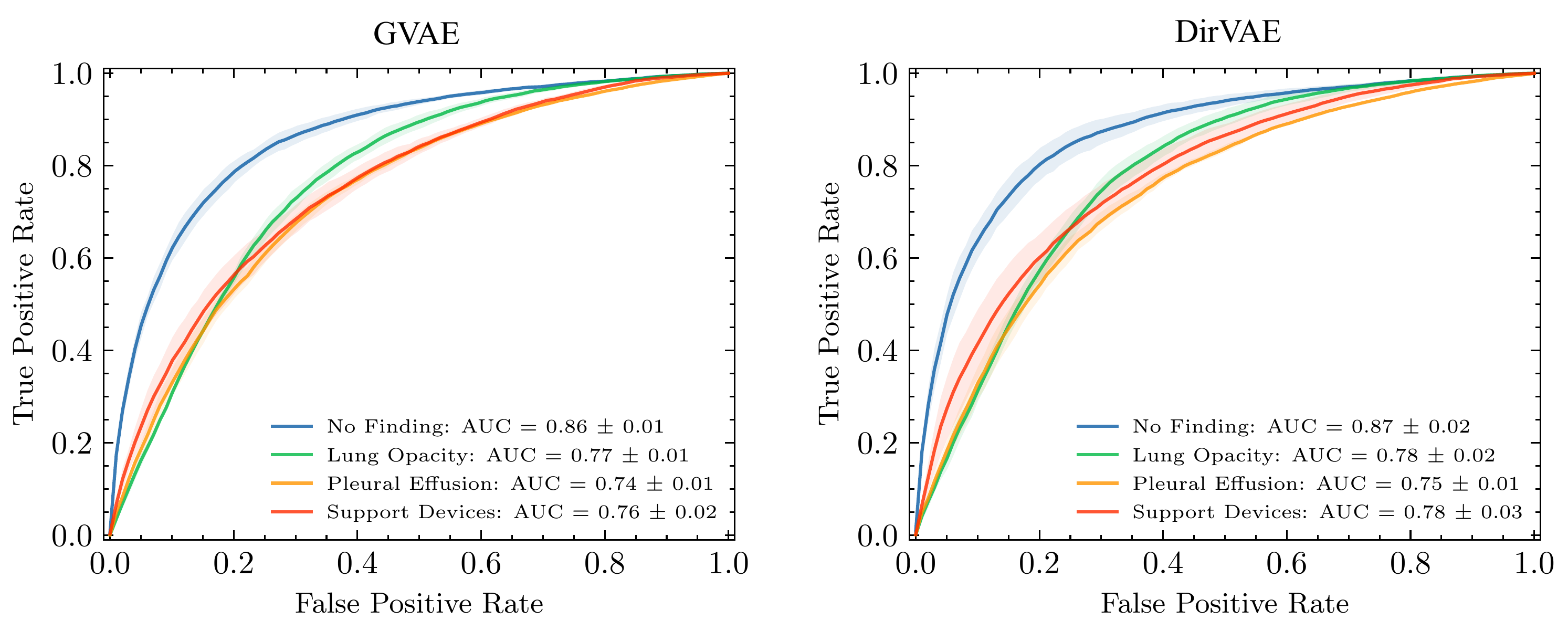}
    \caption{Label-wise ROC curves for GVAE and DirVAE logistic regression classifier performance.}
    \label{fig:roc}
\end{figure}

Table \ref{tab:res} summarises the classification performance of the logistic regression models, for both the DirVAE and the GVAE. The DirVAE classifiers marginally outperform GVAE classifiers with higher values across most classification metrics (see tab. \ref{tab:res}) and higher AUC scores as shown in Figure \ref{fig:roc}.  While individual label prediction metrics are good, multi-label metrics show that the DirVAE classifiers offer only modest performance in the prediction of all four labels, achieving an exact match rate (EMR) of 0.38 $\pm$ 0.01 and Hamming loss of 0.25 $\pm$ 0.02. GVAE achieved an EMR of only 0.37 $\pm$ 0.01 and Hamming loss of 0.26 $\pm$ 0.01. 
\begin{table}[H]
\centering
\begin{tabular}{|c|l|c|c|}
\hline
                                    &                           & \textbf{Dirichlet VAE} & \textbf{Gaussian VAE} \\ \hline
\multirow{4}{*}{\textbf{Accuracy}}  & \textit{No Finding}       & 0.82 $\pm$ 0.01          & 0.82 $\pm$ 0.01       \\ \cline{2-4} 
                                    & \textit{Lung Opacity}     & 0.72 $\pm$ 0.01         & 0.72 $\pm$ 0.01         \\ \cline{2-4} 
                                    & \textit{Pleural Effusion} & 0.70 $\pm$ 0.01          & 0.70  $\pm$ 0.01       \\ \cline{2-4} 
                                    & \textit{Support Devices}  & 0.74 $\pm$ 0.02          & 0.72 $\pm$ 0.01        \\ \hline
\multirow{4}{*}{\textbf{Precision}} & \textit{No Finding}       & 0.61 $\pm$ 0.06          & 0.62 $\pm$ 0.03         \\ \cline{2-4} 
                                    & \textit{Lung Opacity}     & 0.82 $\pm$ 0.06          & 0.82  $\pm$ 0.02      \\ \cline{2-4} 
                                    & \textit{Pleural Effusion} & 0.81 $\pm$ 0.01          & 0.82 $\pm$ 0.02         \\ \cline{2-4} 
                                    & \textit{Support Devices}  & 0.87 $\pm$ 0.01         & 0.89 $\pm$ 0.01         \\ \hline
\multirow{4}{*}{\textbf{Recall}}    & \textit{No Finding}       & 0.74 $\pm$ 0.01          & 0.72 $\pm$ 0.01         \\ \cline{2-4} 
                                    & \textit{Lung Opacity}     & 0.69 $\pm$ 0.02         & 0.68 $\pm$ 0.02        \\ \cline{2-4} 
                                    & \textit{Pleural Effusion} & 0.72 $\pm$ 0.01         & 0.72 $\pm$  0.01         \\ \cline{2-4} 
                                    & \textit{Support Devices}  & 0.76 $\pm$ 0.02         & 0.74 $\pm$ 0.01         \\ \hline
\multirow{4}{*}{\textbf{AUC}}       & \textit{No Finding}       & 0.87  $\pm$ 0.02        & 0.87  $\pm$ 0.01         \\ \cline{2-4} 
                                    & \textit{Lung Opacity}     & 0.78 $\pm$ 0.02        & 0.77 $\pm$ 0.01         \\ \cline{2-4} 
                                    & \textit{Pleural Effusion} & 0.75 $\pm$ 0.01        & 0.74 $\pm$ 0.01         \\ \cline{2-4} 
                                    & \textit{Support Devices}  & 0.78 $\pm$ 0.03          & 0.76 $\pm$ 0.02         \\ \hline
\end{tabular}
\vspace{0.3cm}
\caption{Label-wise table of logistic regression classification performance results}
\label{tab:res}
\end{table}

Figure \ref{fig:dir_trav} presents the results of traversing the latent spaces in the direction of the most predictive latent dimension. For the DirVAE, clear structural changes are observed for each class. Typically, latent traversals will gradually remove or intensify disease-related features, in a manner specific to the input CXR image. We consider both types of change as evidence of disease-/class-specific and image-specific disentanglement and generate pixel intensity variance maps (presented along side each traversal, bottom right corner of each panel) in order to capture both types of feature change. We find that the features highlighted by DirVAE traversals are not only disease- and image-specific, but also clinically relevant. The feature changes observed during latent traversals for the `Pleural Effusion' class are largely isolated to the lower regions of the lungs, typically affecting the appearance of the costophrenic angles and hemidiaphragm region, two keys areas of diagnostic relevance for pleural effusion (fig. \ref{fig:pe_1z}) \cite{pe_radio}.

\begin{figure}
    \centering
    \includegraphics[width=\textwidth]{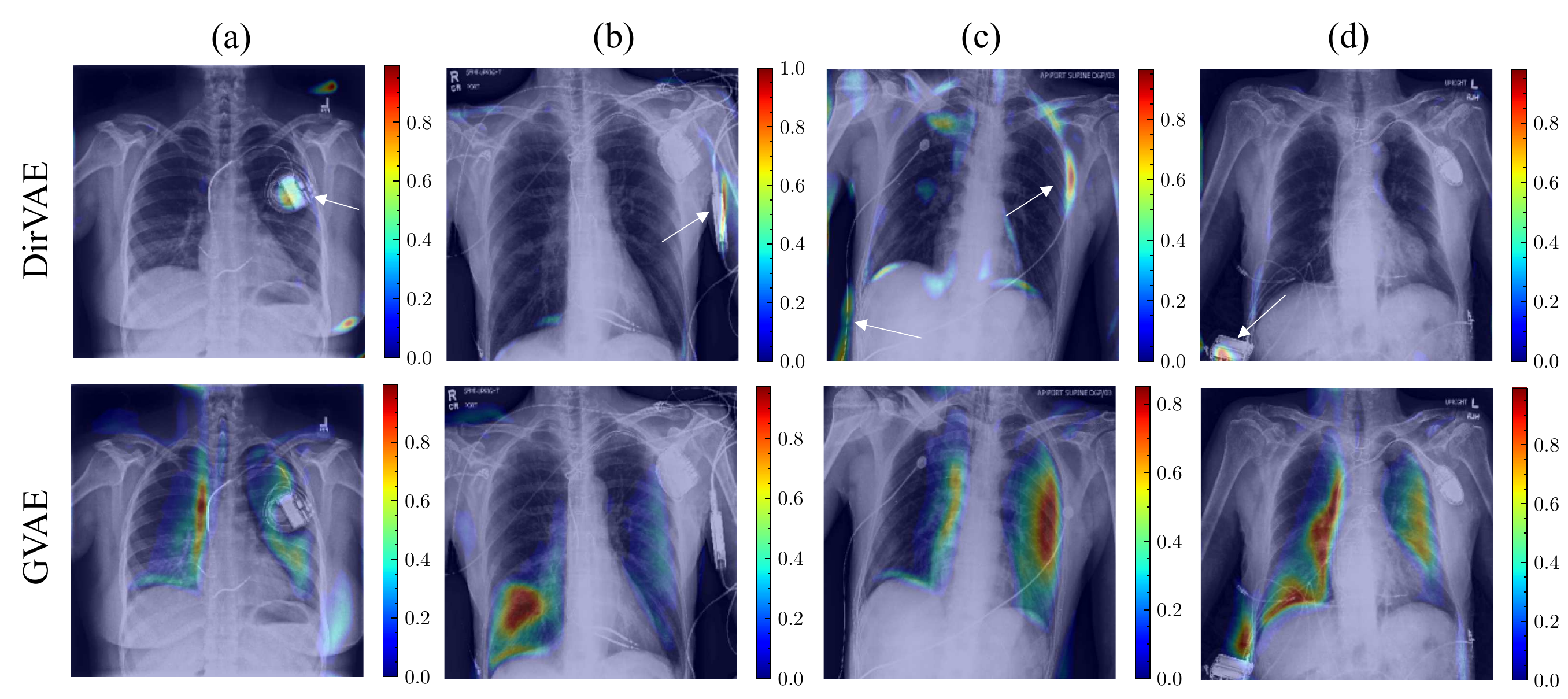}
    \caption{Saliency maps representing variance over all reconstructions in a latent traversal series, performed for the `Support Devices' class, with comparison between GVAE and DirVAE. We include arrows in DirVAE examples to show areas of class-related feature changes.}
    \label{fig:hm_sd}
\end{figure}

For lung opacity cases, DirVAE traversals consistently highlight areas of suspected consolidation. Clinically, the radiographic presentation of lung opacities can vary greatly, this is reflected in the diversity of feature changes observed during DirVAE traversals. Traversals capture features of consolidation that vary in shape, size and position, matching the appearance of diagnostic features in the example CXRs (fig. \ref{fig:lo_1z}). Similarly, in evaluating traversals of the `Support Devices' class, we observe a diverse set of well-defined structural changes in all areas of the CXR, with significant changes mirroring the location and shape of support devices in the evaluated CXR (see fig. \ref{fig:sd_1z} and fig. \ref{fig:hm_sd}). These results indicate that the DirVAE has the capacity to learn disease/class-specific latent factors that are representative of all modes within the class of interest. 

With successful isolation of salient features in the latent space, potential reliance on `spurious' features for class prediction can be observed in some cases (indicated by red boxes overlaid on CXR images in fig. \ref{fig:dir_trav}). Here, we see that latent traversal causes feature changes relating to radiology annotation, co-occurring classes (particularly support devices) and shoulder position. This indicates that the disentangled latent space learned by the DirVAE could also be used to identify spurious features resulting from biases in the data, which is a requisite for mitigating the same. %

\begin{figure}[H]
    \centering
    \includegraphics[width=0.9\textwidth]{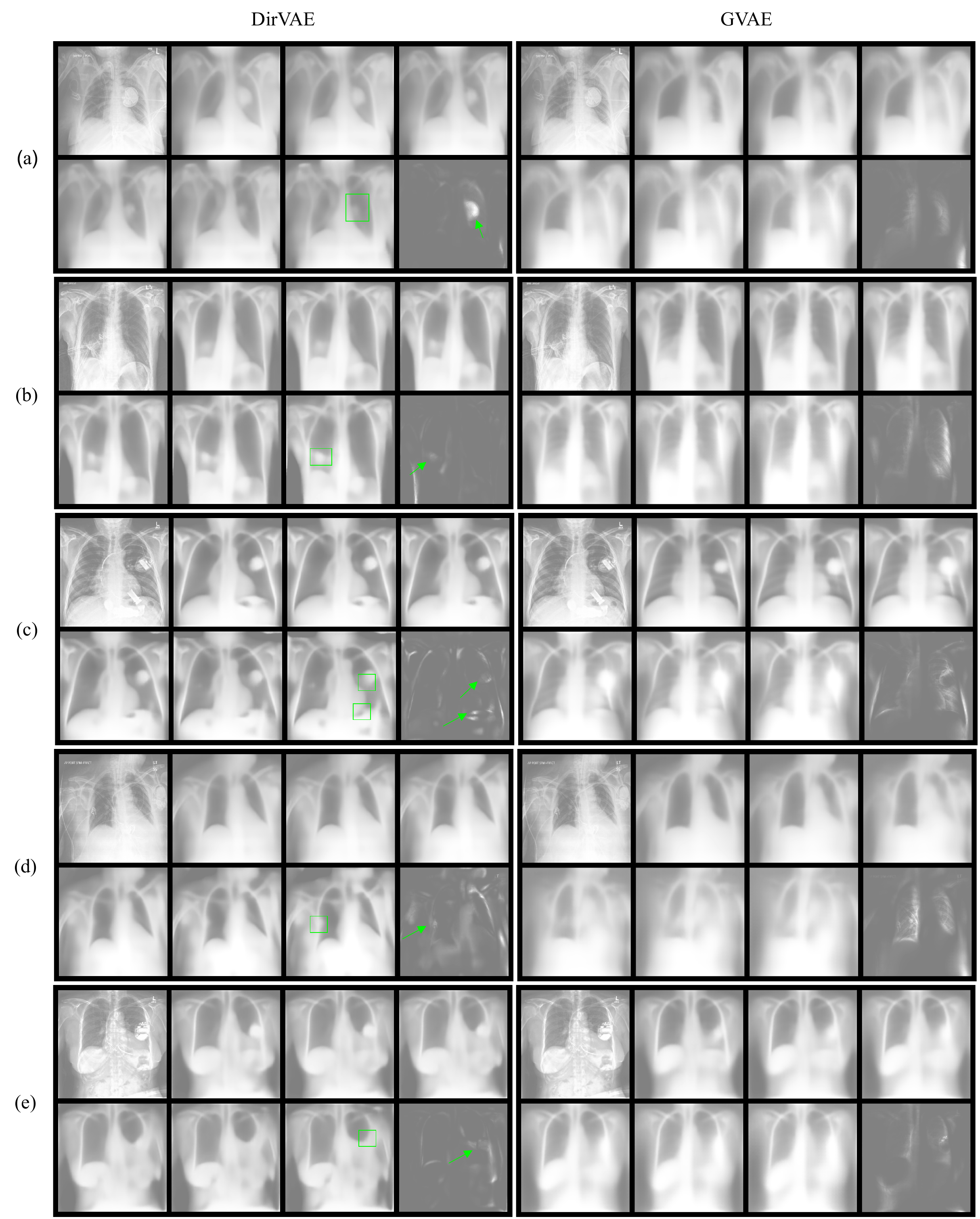}
    \vspace{0.15cm}
    \caption{A direct comparison of GVAE and DirVAE traversals, with examples from the `Support Devices' class. Each row presents a different example, the top left image in the grid is the input CXR, the next image is the original reconstruction, the entire traversal is then presented sequentially in the grid. The bottom right image presents a pixel-wise variance map summarising the entire traversal. Green arrows (\greenright) in the variance map point to class-related feature changes. Similarly, green boxes are used to highlight key features changes in the final reconstruction.}
    \label{fig:dirvsgau}
\end{figure}

Crucially, no such class-specific changes are observed during evaluation of GVAE latent traversals, for any class (fig. \ref{fig:gau_trav}). Traversal variance appears diffuse across the reconstructed image, showing only subtle changes in anatomical areas and little to no change in areas of diagnostic relevance. Where disease-related feature changes are observed they are not localised to regions relevant to the class of interest, but are accompanied by changes to other features across the images. Changes in clavicle, shoulder position, mediastinum and lung width are often observed together, causing non-specific changes in the size or shape of the lung airspace. With non-specific changes GVAE traversals appear similar for all examples and localisation of CXR features of significance is near impossible. This is especially apparent in Figure \ref{fig:dirvsgau} and \ref{fig:hm_sd}, which present direct comparisons between DirVAE and GVAE traversals (for the same input CXR images), with `Support Devices' as the class of interest.

\begin{figure}[H]
\vspace{1cm}
        \centering
            \begin{subfigure}{\textwidth}
            \centering
                \includegraphics[width=0.98\textwidth]{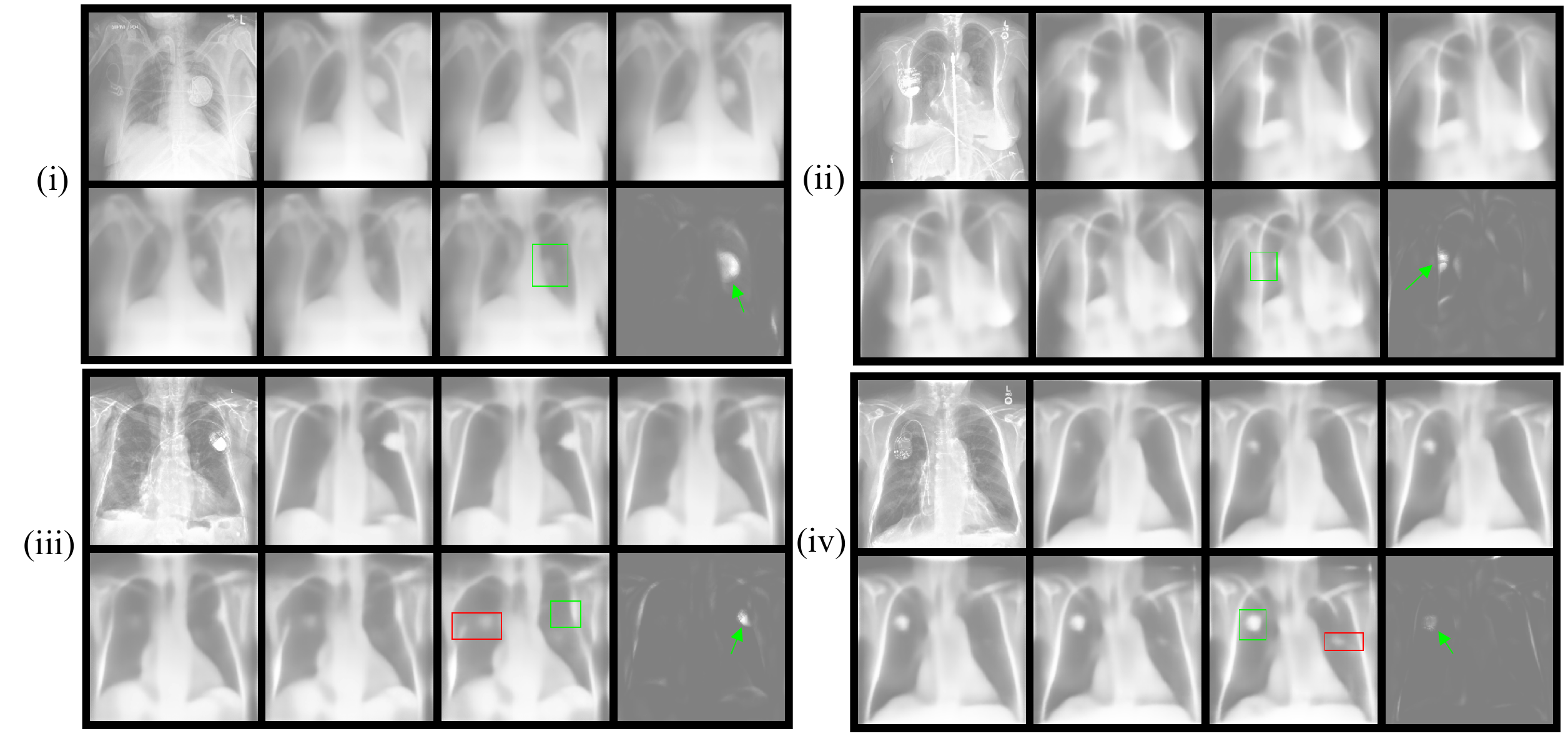}%
                \caption{Examples of DirVAE latent traversal results of the `Support Devices' CXR class}
                \label{fig:sd_1z}
            \end{subfigure}

        \begin{subfigure}{\textwidth}
        \centering
        \vspace{0.5cm}
                \includegraphics[width=0.98\textwidth]{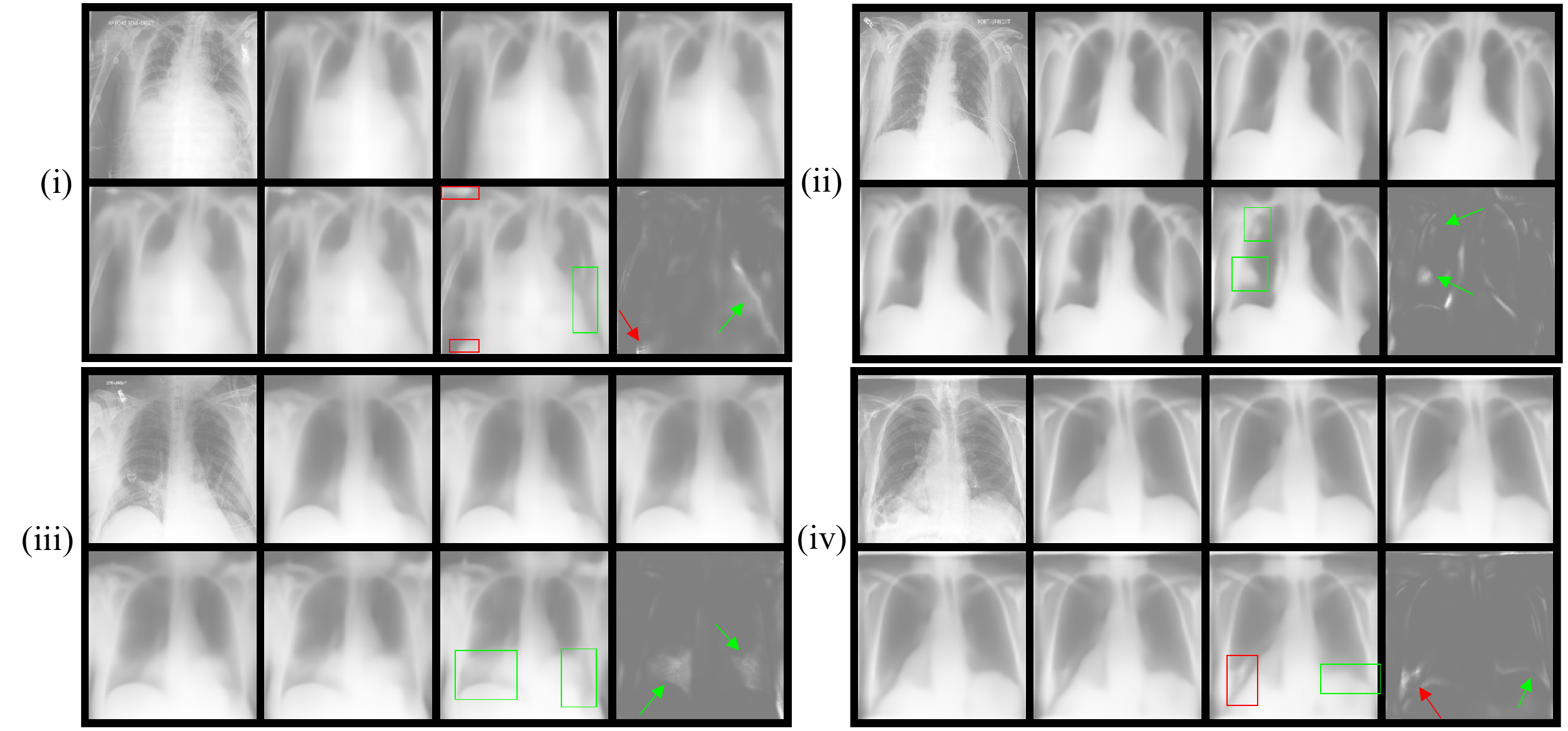}%
                \caption{Examples of DirVAE latent traversal results of the `Lung Opacity' CXR class}
                \label{fig:lo_1z}
        \end{subfigure}
\end{figure}

\begin{figure}[H]
\ContinuedFloat
\centering
        \begin{subfigure}{\textwidth}
        \centering
            \includegraphics[width=0.99\textwidth]{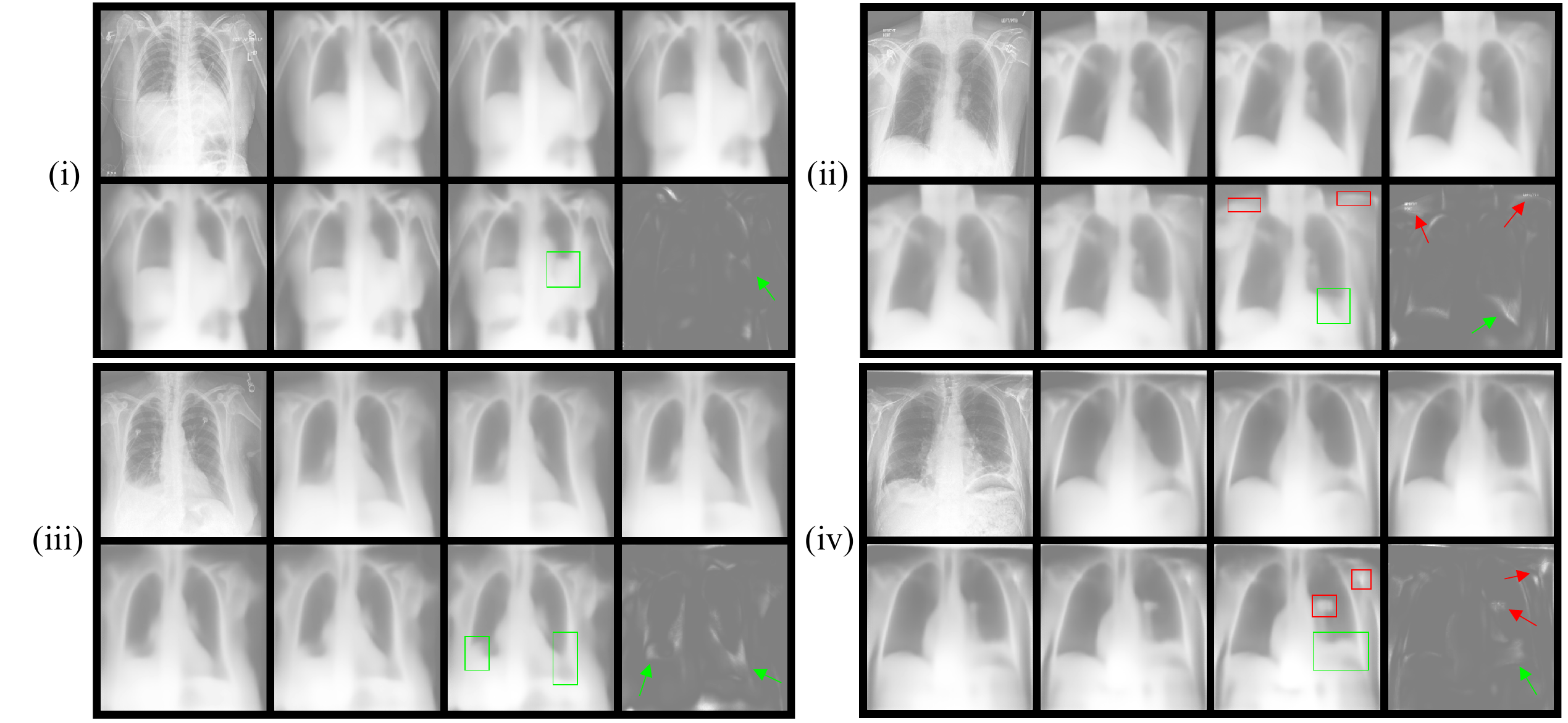} %
            \caption{Examples of DirVAE latent traversal results of the `Pleural Effusion' class}
            \label{fig:pe_1z} 
        \end{subfigure}
        \vspace{0.5cm}
        \caption{DirVAE latent traversals, with examples from each CXR class. Each grid presents a different example, the top left image in the grid is the original reconstruction, the entire traversal is presented sequentially in the grid. The bottom right image presents a pixel-wise variance map summarising the entire traversal. Green arrows (\greenright) in the variance map point to disease-related feature changes while red arrows (\reddright) point to likely confounding features. Similarly, red and green boxes are used to highlight key features changes in the final reconstruction.}
        \label{fig:dir_trav}
\end{figure}

\begin{figure}[H]
\begin{subfigure}{\textwidth}
    \centering
    \includegraphics[width=\textwidth]{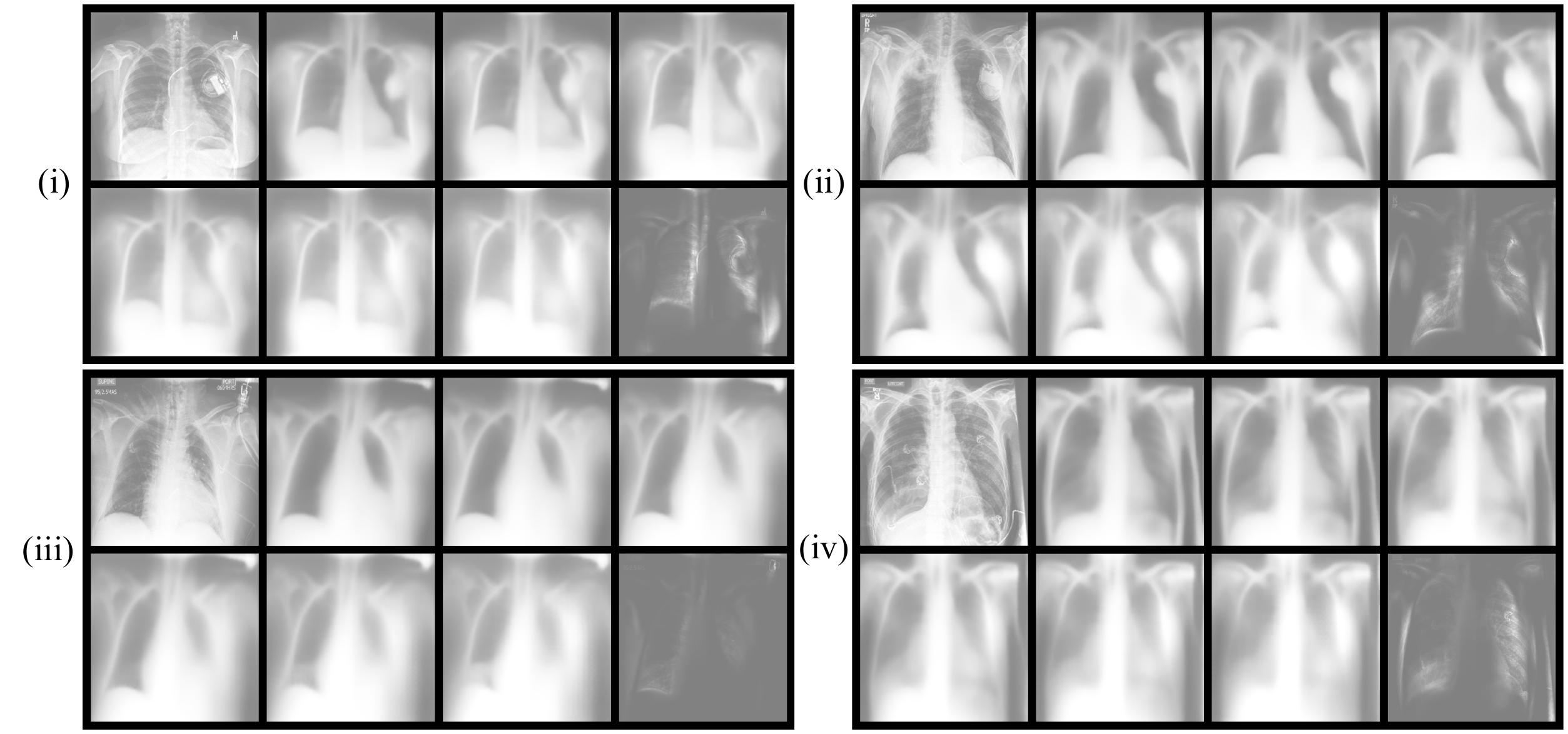} %
    \caption{Examples of GVAE latent traversal results of the ‘Support Devices’ class}
    \label{fig:gau_sd_trav}
 
\end{subfigure}
\end{figure}

\begin{figure}[H]
    \centering
    \ContinuedFloat

    \begin{subfigure}{\textwidth}
    \centering
    \includegraphics[width=\textwidth]{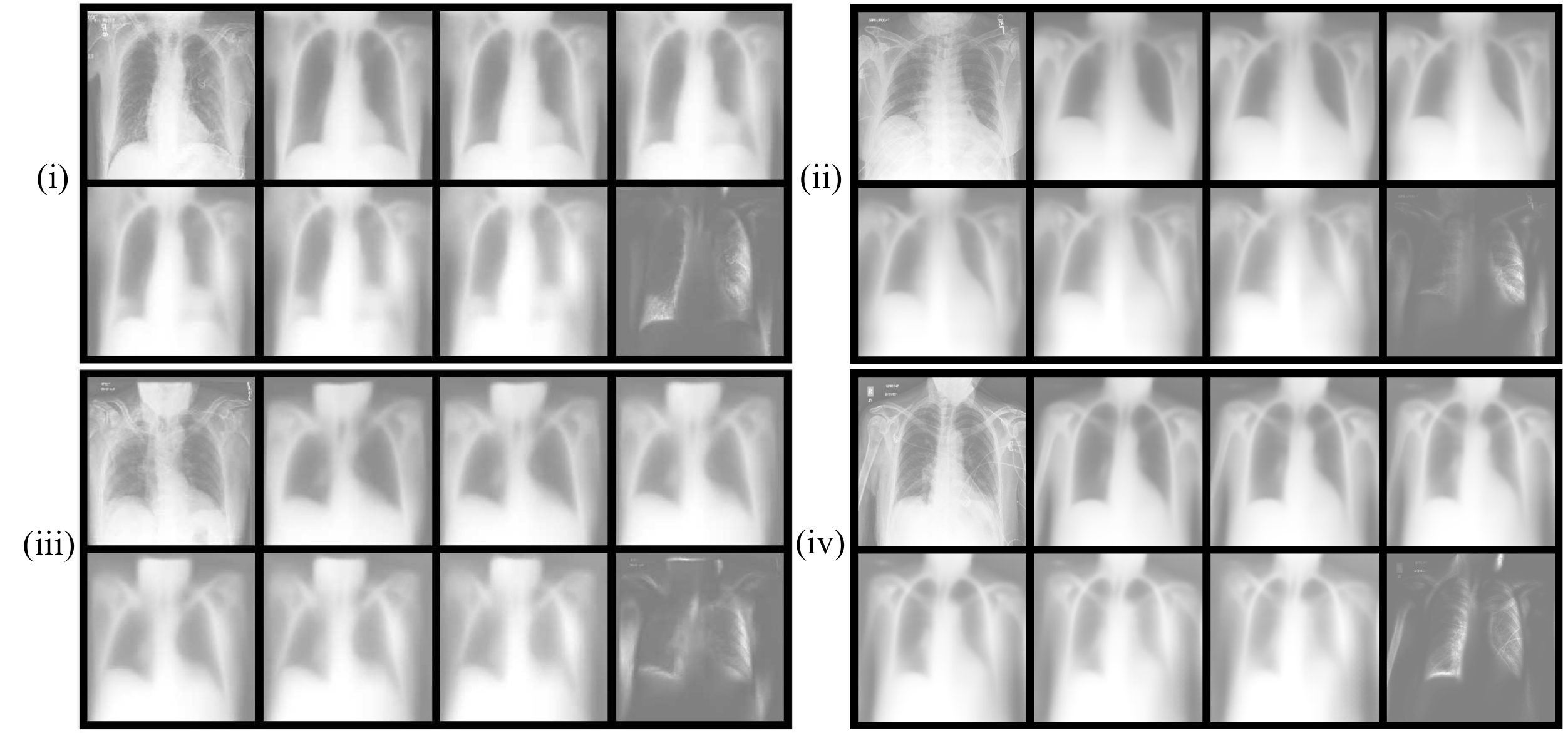} %
    \caption{Examples of GVAE latent traversal results of the `Lung Opacity’ class}
    \label{fig:gau_lo_trav}
\end{subfigure}

\begin{subfigure}{\textwidth}
    \vspace{0.5cm}
    \includegraphics[width=\textwidth]{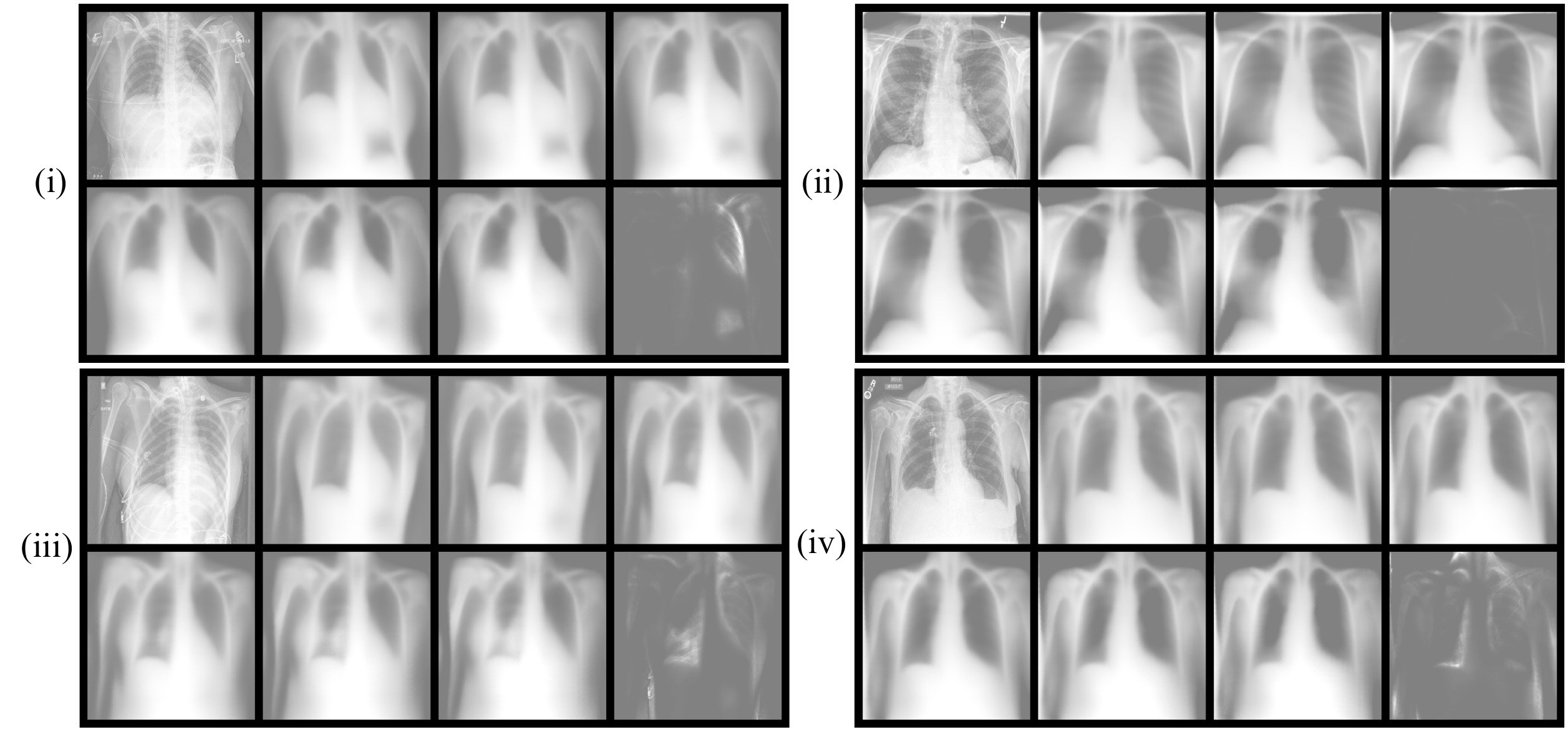}%
    \caption{Examples of GVAE latent traversal results of the ‘Pleural Effusion’ class}
    \label{fig:first}
\end{subfigure}
\label{fig:gau_pe_trav}

\vspace{0.5cm}
\caption{GVAE latent traversals, with examples from each CXR class. Each grid presents a different example, the top left image in the grid is the input CXR, the next image is the original reconstruction, the entire traversal is then presented sequentially in the grid. The bottom right image presents a pixel-wise variance map summarising the entire traversal.}
\label{fig:gau_trav}
\end{figure}

\section{DISCUSSION}

Analysis of latent traversals suggests that the DirVAE learns a more disentangled representation (than its GVAE counterpart), despite the challenges presented by a multi-label task, where the co-occurrence of disease features complicates representation learning. In addition to its superiority in terms of disentanglement and therefore, explainability, DirVAE classifiers perform marginally better than GVAE. The improvement in disentanglement afforded by the DirVAE, relative to the GVAE, is attributed to the sparsity and multi-modal characteristics of the latent space learned by the former. Specifically, the sparsity of our Dirichlet prior is enforced by setting its concentration parameter $\alpha = 0.5$ across all experiments. This pushes areas of the posterior distribution towards zero, i.e. the inner areas of the simplex, and creates a narrower information bottleneck. %

In order to generate quality reconstructions under strict sparsity constraints the latent factors have to capture more global features \cite{ksparse}. Our approach to encouraging disentanglement follows a similar principle to $\beta$-VAE, but with a different strategy for narrowing the information bottleneck \cite{understandingbeta}. If we consider this with our classification objective, which influences the model to prioritise learning of disease-related features, we could say that our strategy encourages the learned latent variables to capture `global disease features' and explain away noise. We can observe this principle in latent traversal evaluations, by traversing latent dimensions important for lung opacity classification, we do not observe diffuse changes in unrelated CXR features, but clear changes in homologous features of the image. Moreover, we observe the capacity of the DirVAE to disentangle global patterns of disease features, with localised regions of consolidation changing across both lungs and in upper and lower lung regions. Similarly, global features are observed in `Support Device' classification, where all support structures, including connecting wiring (distributed spatially across the image) are altered during traversal; during traversal coordinated feature changes can be observed across the entire image, for all images classes. 

Our proposed use of latent embedding manipulation for identification of the salient visual features underlying classifier predictions facilitates improved model explainability. Within our explainability approach, superior disentanglement in DirVAE latent representations allow us to more clearly observe the appearance of `spurious correlations’ alongside disease features, across traversals. Examples of observed `spurious correlation' includes, sex changes, the appearance of various support devices, as well as changing body position (relating to switches from AP to PA projection) \cite{cxr_shortcuts}.

Highlighting the influence of confounders and biases in predictions made by learning-based systems is essential for building safer and fairer predictive models. This is especially relevant for translating learning-based computer-aided diagnostic/screening systems to routine clinical care \cite{Gaube2023}. Particularly, multi-label classification tasks, where the co-occurrence of disease makes deep learning models more vulnerable to reliance on confounding factors. Typical approaches to multi-label image classification explainability, including GradCAM \cite{gradcam}, have been criticised for their inability to highlight smaller pathologies or structures with complex shapes, for example, mechanical wiring \cite{saporta2022benchmarking}. Our work opens doors to a novel approach for this task. By interrogating the direct influence of important/identified latent factors on the generated images and the classification of independent outcomes (or indeed the fluctuation in the predicted class-specific probabilities) for the generated images, our approach provides a systematic framework to identify relevant class-specific features and confounders and/or biases in image data. Interrogating the effects of the most predictive latent factor demonstrates that this approach is able to clearly highlight small and complex features as well as important correlations with, potentially, confounding image features \cite{saporta2022benchmarking}. This is well-illustrated by an example of confounding highlighted in a `Lung Opacity' traversal (fig. \ref{fig:lo_1z} (iv)), where the classifier appears to mistake the gas in the digestive system for opacities in the lower lung regions (highlighted by red box and arrows).

Strong classification results alongside unclear traversals suggest that GVAE classifiers rely on latent factors that combine to explain shared variation and describe the CXR image (and its corresponding class(es)) as a whole, rather than rely on a sparse set of disentangled latent factors that describe the presence/absence of disease-/class-specific features (as with the DirVAE). While some disease-/class-specific features are visible in the latent traversals visualised for GVAE, these are attributed to the effects of multi-task representation learning, i.e. to the joint training of the GVAE with the logistic regression classifiers for the multi-label classification problem. Due to the dense and unimodal nature of Gaussian distributions, the learned latent space appears unable to separate disease-/class-specific features other image-specific features. This is particularly evident in the latent traversals presented for CXR images from the `Support Devices' class (see fig. \ref{fig:dirvsgau} and fig. \ref{fig:hm_sd}). From this, we can conclude that GVAE has limited functionality as a method for prediction explainability.

\section{CONCLUSION}
In this work we demonstrate that disentanglement of class-/disease-specific (and potentially, clinically-relevant) features can be achieved using our DirVAE model and that its capacity for class-/disease-specific disentanglement is superior to a GVAE. We introduce a promising new approach for explainable multi-label classification, where we apply an ensemble of simple logistic regression classifiers and explore latent dimensions of significance to class predictions through so-called `latent traversals'. With this we provide visual examples of how our explainability method can highlight regions in CXR images clinically relevant to the class(es) of interest and additionally, identify cases where classification is biased and relies on spurious feature correlations.

\section{FUTURE WORK}
Future work will explore the use of metrics to quantify disentanglement in order to formalise our qualitative assessment. We will improve image reconstruction/generation quality, with a view to improve the clarity of feature changes during latent traversals. %
We will explore latent space interpolations between CXR images of different classes and observe corresponding changes to classifier predictions during the transitions between disease representations, for further explainability. With improved reconstruction quality the developed approach would have applications in both explainable medical AI and synthetic data.

\section*{ACKNOWLEDGEMENTS}
This study was partially supported by cloud computing credits awarded by Amazon Web Services through the Diagnostic Development Initiative. Kieran Zucker is funded by the NIHR for this research project. The views expressed in this publication are those of the author(s) and not necessarily those of the NIHR, NHS or the UK Department of Health and Social Care.

\bibliography{report} %
\bibliographystyle{spiebib} %

\end{document}